\documentclass[runningheads]{llncs}

\usepackage[paperheight=233mm,paperwidth=152mm,%
  textwidth=12.2cm,  
  textheight=19.3cm, 
  hratio=1:1,      
  vratio=1:1,      
  ]{geometry}

\usepackage{amsmath}
\usepackage{amssymb}
\usepackage{xcolor}
\usepackage{graphicx}
\usepackage{array}
\usepackage{multirow}
\usepackage{makecell}
\usepackage{xspace}
\usepackage[citecolor=blue, urlcolor=blue, colorlinks=true]{hyperref}
\usepackage{cleveref}
\usepackage[nolist]{acronym}
\usepackage{listings}
\usepackage{lipsum}
\usepackage{booktabs}
\usepackage{footnote}
\usepackage{bm} 
\usepackage{placeins}
\usepackage{enumitem}
\usepackage{emptypage}
\usepackage{imakeidx}
    \makeindex
    \makeindex[name=authors,title=Authors,columns=2]

\newcommand{\thetitle}{\arabic{title}}
\newcounter{title}

\newcommand{\mytitle}[1]{
	\stepcounter{title}
	\title{#1}
	\titlerunning{\thetitle.\space#1}
	\toctitle{\thetitle.\space#1}
}

\newcommand{\preftitle}[1]{
	\title{Preface}
	\titlerunning{}
	\toctitle{}
}

\renewcommand{\thesection}{\arabic{section}}

\renewcommand{\thefigure}{\thetitle.\arabic{figure}}
\renewcommand{\thetable}{\thetitle.\arabic{table}}

\renewcommand{\appendix}{%
\setcounter{section}{0}
\renewcommand{\thesection}{\thetitle.\Alph{section}}
}

\renewcommand\thanks[1]{%
  \begingroup
  \renewcommand\thefootnote{}\footnote{\hskip -3mm #1}%
  \addtocounter{footnote}{-1}%
  \endgroup
}

\begin{document}

\mytitle{Towards Explainable Artificial Intelligence}
\author{Wojciech Samek\inst{1} \and Klaus-Robert M\"uller\inst{2,3,4}\thanks{The final authenticated publication is available online at \url{https://doi.org/10.1007/978-3-030-28954-6\_1}. In: W. Samek et al. (Eds.) Explainable AI: Interpreting, Explaining and Visualizing Deep Learning. Lecture Notes in Computer Science, vol. 11700, pp. 5-22. Springer, Cham (2019).}}
\authorrunning{W. Samek and K.-R. M\"uller}
\institute{Fraunhofer Heinrich Hertz Institute, 10587 Berlin, Germany
\email{wojciech.samek@hhi.fraunhofer.de} \and
Technische Universit\"at Berlin, 10587 Berlin, Germany \and
Korea University, Anam-dong, Seongbuk-gu, Seoul 02841, Korea \and
Max Planck Institute for Informatics, Saarbr\"ucken 66123, Germany\\
\email{klaus-robert.mueller@tu-berlin.de}}
\maketitle

\begingroup

\begin{abstract}
In recent years, machine learning (ML) has become a key enabling technology for the sciences and industry.
Especially through improvements in methodology, the availability of large databases and increased computational power, today's ML algorithms are able to achieve excellent performance (at times even exceeding the human level) on an increasing number of complex tasks. Deep learning models are at the forefront of this development. However, due to their nested non-linear structure, these powerful models have been generally considered ``black boxes'',\index{Black-Box AI} not providing any information about what exactly makes them arrive at their predictions. 
Since in many applications, e.g., in the medical domain, such lack of transparency may be not acceptable, the development of methods for visualizing, explaining and interpreting deep learning models has recently attracted increasing attention. This introductory paper presents recent developments and applications in this field and makes a plea for a wider use of {\it explainable} learning algorithms in practice. 
\index[authors]{Samek, Wojciech}
\index[authors]{M{\"u}ller, Klaus-Robert}
\end{abstract}
%
%
\section{Introduction}
\label{samek:sec:intro}
Today's artificial intelligence (AI) systems based on machine learning excel in many fields.
They not only outperform humans in complex visual tasks \cite{samek:cirecsan2011committee,samek:lu2014surpassing} or strategic games \cite{samek:mnih2015human,samek:silver2017mastering,samek:MorScience17}, but also became an indispensable part of our every day lives, e.g., as intelligent cell phone cameras which can recognize and track faces \cite{samek:redmon2016you}, as online services which can analyze and translate written texts \cite{samek:bahdanau2014neural} or as consumer devices which can understand speech and generate human-like answers \cite{samek:van2016wavenet}. Moreover, machine learning and artificial intelligence have  become indispensable tools in the sciences for tasks such as prediction, simulation or exploration \cite{samek:schutt2017quantum,samek:Chmiela2018,samek:ThoArXiv18,samek:wu2019solving}. These immense successes of AI systems mainly became possible through improvements in deep learning methodology \cite{samek:lecun2012efficient,samek:lecun2015deep}, the availability of large databases \cite{samek:deng2009imagenet,samek:karpathy2014large}  and computational gains obtained with powerful GPU cards \cite{samek:lindholm2008nvidia}. 

Despite the revolutionary character of this technology, challenges still exist which slow down or even hinder the prevailance of AI in some applications.
Examplar challenges are (1) the large complexity and high energy demands of current deep learning models \cite{samek:han2015learning}, which hinder their deployment in resource restricted environments  and devices, (2) the lack of robustness to adversarial attacks \cite{samek:madry2017towards}, which pose a severe security risk in application such as autonomous driving\footnote{The authors of \cite{samek:eykholt2017robust} showed that deep models can be easily fooled by physical-world attacks. For instance, by putting specific stickers on a stop sign one can achieve that the stop sign is not recognized by the system anymore.}, and (3) the lack of transparency and explainability \cite{samek:SamITU18b,samek:holzinger2019causability,samek:doshi2017towards}, which reduces the trust in and the verifiability of the decisions made by an AI system.

This paper focuses on the last challenge. It presents recent developments in the field of {\it explainable artificial intelligence}\index{Explainable AI}\index{XAI (Explainable AI)} and aims to foster awareness for the advantages--and at times--also for the necessity of transparent decision making in practice. The historic second Go match between Lee Sedol and AlphaGo \cite{samek:silver2016mastering} nicely demonstrates the power of today's AI technology, and hints at its enormous potential for generating new knowledge from data when being accessible for human interpretation. In this match AlphaGo played a move, which was classified as ``not a human move'' by a renowned Go expert, but which was the deciding move for AlphaGo to win the game. AlphaGo did not explain the move, but the later play unveiled the intention behind its decision.
With explainable AI it may be possible to also identify such novel patterns and strategies in domains like health, drug development or material sciences, moreover, the explanations will ideally let us comprehend the reasoning of the system and understand why the system has decided e.g.\ to classify a patient in a specific manner or associate certain properties with a new drug or material. This opens up innumerable possibilities for future research and may lead to new scientific insights. 

The remainder of the  paper is organized as follows. Section \ref{samek:sec:xai} discusses the need for transparency and trust in AI.
Section \ref{samek:sec:explain} comments on the different types of explanations and their respective information content and use in practice.
Recent techniques of explainable AI are briefly summarized in Section \ref{samek:sec:methods}, including methods which rely on simple surrogate functions, frame explanation as an optimization problem, access the model's gradient or make use of the model's internal structure. The question of how to objectively evaluate the quality of explanations is addressed in Section \ref{samek:sec:quality}. The paper concludes in Section \ref{samek:sec:challenges} with a discussion on general challenges in the field of explainable AI.

\section{Need for Transparency and Trust in AI}
\label{samek:sec:xai}
\index{Explainable AI!Motivations}%
\index{Trust}%
\index{Transparency}
Black box AI systems have spread to many of today's applications. For machine learning models used, e.g., in consumer electronics or online translation services, transparency and explainability are not a key requirement as long as the overall performance of these systems is good enough. But even if these systems fail, e.g., the cell phone camera does not recognize a person or the translation service produces grammatically wrong sentences, the consequences are rather unspectacular. Thus, the requirements for transparency and trust are rather low for these types of AI systems.
In safety critical applications the situation is very different. Here, the intransparency of ML techniques may be a limiting or even disqualifying factor. Especially if single wrong decisions can result in danger to life and health of humans (e.g., autonomous driving, medical domain) or significant monetary losses (e.g., algorithmic trading), relying on a data-driven system whose reasoning is incomprehensible may not be an option. 
This intransparency is one reason why the adoption of machine learning to domains such as health is more cautious than the usage of these models in the consumer, e-commerce or entertainment industry.

In the following we discuss why the ability to explain the decision making of an AI system helps to establish trust and is of utmost importance, not only in medical or safety critical applications. We refer the reader to \cite{samek:weller2019} for a discussion of the challenges of transparency.

\subsection{Explanations Help to Find ``Clever Hans'' Predictors}
\index{Clever Hans Predictor}%
\index{Validating ML with Explanations!Clever Hans Predictors}
Clever Hans was a horse that could supposedly count and that was considered a scientific sensation in the years around 1900. As it turned out later, Hans did not master the math but in about 90 percent of the cases, he was able to derive the correct answer from the questioner's reaction. Analogous behaviours have been recently observed in state-of-the-art AI systems \cite{samek:LapNCOMM19}. Also here the algorithms have learned to use some spurious correlates in the training and test data and similarly to Hans predict right for the `wrong' reason.

For instance, the authors of \cite{samek:LapCVPR16,samek:LapNCOMM19} showed that the winning method of the PASCAL VOC competition \cite{samek:everingham2010pascal} was often not detecting the object of interest, but was utilizing correlations or context in the  data to correctly classify an image. It recognized boats by the presence of water and trains by the presence of rails in the image, moreover, it recognized horses by the presence of a copyright watermark\footnote{The PASCAL VOC images have been automatically crawled from flickr and especially the horse images were very often copyrighted with a watermark.}. The occurrence of the copyright tags in  horse images is a clear artifact in the dataset, which had gone unnoticed to the organizers and participants of the challenge for many years. It can be assumed that nobody has systematically checked the thousands images in the dataset for this kind of artifacts (but even if someone did, such artifacts may be easily overlooked). Many other examples of ``Clever Hans'' predictors have been described in the literature. For instance, \cite{samek:ribeiro2016should} show that current deep neural networks are distinguishing the classes ``Wolf'' and ``Husky'' mainly by the presence of snow in the image. The authors of \cite{samek:LapNCOMM19}  demonstrate that deep models overfit to padding artifacts when classifying airplanes, whereas \cite{samek:mordvintsev2015inceptionism} show that a model which was trained to distinguish between 1000 categories, has not learned dumbbells as an independent concept, but associates a dumbbell with the arm which lifts it.
Such ``Clever Hans'' predictors perform well on their respective test sets, but will certainly fail if deployed to the real-world, where sailing boats may lie on a boat trailer, both wolves and huskies can be found in non-snow regions  and horses do not have a copyright sign on them. 
However, if the AI system is a black box, it is very difficult to unmask such predictors.
Explainability helps to detect these types of biases in the model or the data, moreover, it helps to understand the weaknesses of the AI system (even if it is not a ``Clever Hans'' predictor). In the extreme case, explanations allow to detect the classifier's misbehaviour (e.g., the focus on the copyright tag) from a single test image\footnote{Traditional methods to evaluate  classifier performance require large test datasets.}. Since understanding the weaknesses of a system is the first step towards improving it, explanations are likely to become integral part of the training and validation process of future AI models.

\subsection{Explanations Foster Trust and Verifiability}
\index{Uses of Explanation!Debugging}%
\index{Validating ML with Explanations!Trust}%
\index{Societal Aspects!Trust}%
\index{Societal Aspects!User Acceptance}%
\index{Validating ML with Explanations!Verifiability}
The ability to verify decisions of an AI system is very important to foster trust, both in situations where the AI system has a supportive role (e.g., medical diagnosis) and in situations where it practically takes the decisions (e.g., autonomous driving). In the former case, explanations provide extra information, which, e.g., help the medical expert to gain a comprehensive picture of the patient in order to take the best therapy decision. Similarly to a radiologist, who writes a detailed report explaining his findings, a supportive AI system should in detail explain its decisions rather than only providing the diagnosis to the medical expert. In cases where the AI system itself is deciding, it is even more critical to be able to comprehend the reasoning of the system in order to verify that it is not behaving like Clever Hans, but solves the problem in a robust and safe manner. Such verifications are required to build the necessary trust in every new technology. 

There is also a social dimension of explanations. Explaining the rationale behind one's decisions is an important part of human interactions \cite{samek:heath2013human}. Explanations help to build trust in a relationship between humans, and should therefore be also part of human-machine interactions \cite{samek:antunes2012structuring}. Explanations are not only an inevitable part of human learning and education (e.g., teacher explains solution to student), but also foster the acceptance of difficult decisions and are important for informed consent (e.g., doctor explaining therapy to patient). Thus, even if not providing additional information for verifying the decision, e.g., because the patient may have no medical knowledge, receiving explanations usually make us feel better as it integrates us into the decision-making process. An AI system which interacts with humans should therefore be explainable.

\subsection{Explanations are a Prerequisite for New Insights}
\index{Uses of Explanation!Build Novel Insights}
AI systems have the potential to discover patterns in data, which are not accessible to the human expert. 
In the case of the Go game, these patterns can be new playing strategies \cite{samek:silver2016mastering}. In the case of scientific data, they can be unknown associations between genes and diseases \cite{samek:libbrecht2015machine}, chemical compounds and material properties \cite{samek:pilania2013accelerating} or brain activations and cognitive states \cite{samek:lemm2011introduction}. In the sciences, identifying these patterns, i.e., explaining and interpreting what features the AI system uses for predicting, is often more important than the prediction itself, because it unveils information about the biological, chemical or neural mechanisms and may lead to new scientific insights. 

This necessity to explain and interpret the results has led to a strong dominance of linear models in scientific communities in the past (e.g.\ \cite{samek:kriegeskorte2006information,samek:phinyomark2018analysis}).
Linear models are intrinsically interpretable and thus easily allow to extract the learned patterns. Only recently, it became possible to apply more powerful models such as deep neural networks without sacrificing interpretability. These explainable non-linear models have already attracted attention in domains such as neuroscience \cite{samek:StuJNM16,samek:ThoArXiv18,samek:eitel2019uncovering}, health \cite{samek:HorArXiv18,samek:binder2018towards,samek:Klauschen2018scoring}, autonomous driving \cite{samek:hofmarcher}, drug design \cite{samek:preuer} and physics \cite{samek:schutt2017quantum,samek:reyes2018enhanced} and it can be expected that they will play a pivotal role in future scientific research.

\subsection{Explanations are Part of the Legislation}
\index{Uses of Explanation!Legal Compliance}
\index{Legal Aspects}
The infiltration of AI systems into our daily lives poses a new challenge for the legislation. Legal and ethical questions regarding the responsibility of AI systems and their level of autonomy have recently received increased attention \cite{samek:eu,samek:goodman2017european}. But also anti-discrimination and fairness aspects have been widely discussed in the context of AI \cite{samek:hajian2016algorithmic,samek:doshi2017accountability}. The  EU's General Data Protection Regulation (GDPR) has even added the {\it right to explanation}\index{Legal Aspects!Right to Explanation}\index{Right to Explanation} to the policy in Articles 13, 14 and 22, highlighting the importance of human-understandable interpretations derived from machine decisions. For instance, if a person is being rejected for a loan by the AI system of a bank, in principle, he or she has the right to know why the system has decided in this way, e.g., in order to make sure that the decision is compatible with the anti-discrimination law or other regulations. Although it is not yet clear how these legal requirements will be implemented in practice, one can be sure that transparency aspects will gain in importance as AI decisions will more and more affect our daily lives.

\section{Different Facets of an Explanation}
\label{samek:sec:explain}
Recently proposed explanation techniques provide valuable information about the learned representations and the decision-making of an AI system.
These explanations may differ in their information content, their recipient and their purpose.\index{Explanations}
In the following we describe the different types of explanations and comment on their usefulness in practice.

\subsection{Recipient}
\index{Explanations!Recipient}
Different recipients may require explanations with different level of detail and with different information content.
For instance, for users of AI technology it may be sufficient to obtain coarse explanations, which are easy to interpret, whereas AI researchers and developers would certainly prefer explanations, which give them deeper insights into the functioning of the model. 

In the case of image classification such simple explanations could coarsely highlight image regions, which are regarded most relevant for the model. Several preprocessing steps, e.g., smoothing, filtering or contrast normalization, could be applied to further improve the visualization quality. 
Although discarding some information, such coarse explanations could help the ordinary user to foster trust in AI technology.
On the other hand AI researchers and developers, who aim to improve the model, may require all the available information, including negative evidence, about the AI's decision in the highest resolution (e.g., pixel-wise explanations), because only this complete information gives detailed insights into the (mal)functioning of the model. 

One can easily identify further groups of recipients, which are interested in different types of explanations. For instance, when applying AI to the medical domain these groups could be patients, doctors and institutions. An AI system which analyzes patient data could provide simple explanations to the patients, e.g., indicating too high blood sugar, while providing more elaborate explanations to the medical personal, e.g., unusual relation between different blood parameters. Furthermore, institutions such as hospitals or the FDA might be less interested in understanding the AI's decisions for individual patients, but would rather prefer to obtain global or aggregated explanations, i.e., patterns which the AI system has learned after analyzing many patients. 

\subsection{Information Content}
\index{Explainable AI!Types of Transparency}
\index{Explanations!Types}
Different types of explanation provide insights into different aspects of the model, ranging from information about the learned representations to the identification of distinct prediction strategies and the assessment of overall model behaviour. Depending on the recipient of the explanations and his or her intent, it may be advantageous to focus on one particular type of explanation. In the following we briefly describe four different types of explanations.

\begin{enumerate}
\item {\bf Explaining learned representations}: 
\index{Types of Explanation!Signal/Prototype}
This type of explanation aims to foster the understanding of the learned representations, e.g., neurons of a deep neural network.
Recent work \cite{samek:bau2017network,samek:kim2017interpretability} investigates the role of single neurons or group of neurons in encoding certain concepts.
Other methods \cite{samek:DBLP:journals/corr/SimonyanVZ13,samek:yosinski2015understanding,samek:nguyen2016synthesizing,samek:nguyen2019} aim to interpret what the model has learned by building prototypes that are representative of the abstract learned concept. These methods, e.g., explain what the model has learned about the category ``car'' by generating a prototypical image of a car. Building such a prototype can be formulated within the activation maximization framework and has been shown to be an effective tool for studying the internal representation of a deep neural network.
\item {\bf Explaining individual predictions}: 
\index{Types of Explanation!Attribution Heatmap}
Other types of explanations provide information about individual predictions, e.g., heatmaps visualizing which pixels have been most relevant for the model to arrive at its decision \cite{samek:montavon2017methods} or heatmaps highlighting the most sensitive parts of an input \cite{samek:DBLP:journals/corr/SimonyanVZ13}. Such explanations help to verify the predictions and establish trust in the correct functioning on the system. Layer-wise Relevance Propagation (LRP) \cite{samek:BachPLOS15,samek:montavon2019} provides a general framework for explaining individual predictions, i.e., it is applicable to various ML models, including neural networks \cite{samek:BachPLOS15}, LSTMs \cite{samek:ArrWASSA17}, Fisher Vector classifiers \cite{samek:LapCVPR16} and Support Vector Machines \cite{samek:kauffmann2018towards}. Section \ref{samek:sec:methods} gives an overview over recently proposed methods for computing  individual explanations.
\item {\bf Explaining model behaviour}: 
\index{Types of Explanation!Meta-Explanation}
This type of explanations go beyond the analysis of individual predictions towards a more general understanding of model behaviour, e.g., identification of distinct prediction strategies. The spectral relevance analysis (SpRAy) approach of \cite{samek:LapNCOMM19} computes such meta explanations by clustering individual heatmaps. Each cluster then represents a particular prediction strategy learned by the model. For instance, the authors of \cite{samek:LapNCOMM19} identify four clusters when classifying ``horse'' images with the Fisher Vector classifier \cite{samek:sanchez2013image} trained on the PASCAL VOC 2007 dataset \cite{samek:everingham2015pascal}, namely (1) detect the horse and rider, 2) detect a copyright tag in portrait oriented images, 3) detect wooden hurdles and other contextual elements of horseback riding, and 4) detect a copyright tag in landscape oriented images.
Such explanations are useful for obtaining a global overview over the learned strategies and detecting ``Clever Hans'' predictors \cite{samek:LapNCOMM19}.
\item {\bf Explaining with representative examples}: 
\index{Types of Explanation!Representative Examples}
Another class of methods interpret classifiers by identifying representative training examples \cite{samek:koh2017understanding,samek:khanna2018interpreting}. This type of explanations can be useful for obtaining a better understanding of the training dataset and how it influences the model. Furthermore, these representative examples can potentially help to identify biases in the data and make the model more robust to variations of the training dataset.
\end{enumerate}

\subsection{Role}
\index{Uses of Explanation}
Besides the recipient and information content it is also important to consider the purpose of an explanation. Here we can distinguish two aspects, namely (1) the intent of the explanation method (what specific question does the explanation answer) and (2) our intent (what do we want to use the explanation for).

Explanations are relative and it makes a huge difference whether their intent is to explain the prediction as is (even if it is incorrect), whether they aim to visualize what the model ``thinks'' about a specific class (e.g., the true class) or whether they explain the prediction relative to another alternative (``why is this image classified as car and not as truck''). 
Methods such as LRP allow to answer all these different questions, moreover, they also allow to adjust the amount of positive and negative evidence in the explanations, i.e., visualize what speaks for (positive evidence) and against (negative evidence) the prediction. Such fine-grained explanations foster the understanding of the classifier and the problem at hand.

Furthermore, there may be different goals for using the explanations beyond visualization and verification of the prediction. For instance, explanations can be potentially used to improve the model, e.g., by regularization \cite{samek:ross2017right}. Also since explanations provide information about the (relevant parts of the) model, they can be potentially used for model compression and pruning. Many other uses (certification of the model, legal use) of explanations can be thought of, but the details remain future work.

\section{Methods of Explainable AI}
\label{samek:sec:methods}
\index{Explainable AI!Methods}
\index{Explanation Methods}
This section gives an overview over different approaches to explainable AI, starting with techniques which are model-agnostic and rely on a simple surrogate function to explain the predictions. Then, we discuss methods which compute explanations by testing the model's response to local perturbations (e.g., by utilizing gradient information or by optimization). Subsequently, we present very efficient propagation-based explanation techniques which leverage the model's internal structure. Finally, we consider methods which go beyond individual explanations towards a meta-explanation of model behaviour. 

This section is not meant to be a complete survey of explanation methods, but it rather summarizes the most important developments in this field. Some approaches to explainable AI, e.g., methods which find influencial examples \cite{samek:khanna2018interpreting}, are not discussed in this section.

\subsection{Explaining with Surrogates}
\index{Explanation Methods!Surrogate Models}
\index{Surrogate-Based Explanations}
Simple classifiers such as linear models or shallow decision trees are intrinsically interpretable, so that explaining its predictions becomes a trivial task.
Complex classifiers such as deep neural networks or recurrent models on the other hand contain several layers of non-linear transformations, which largely complicates the task of finding what exactly makes them arrive at their predictions. 

One approach to explain the predictions of complex models is to locally approximate them with a simple surrogate function, which is interpretable.
A popular technique falling into this category is Local Interpretable Model-agnostic Explanations (LIME) \cite{samek:ribeiro2016should}.%
\index{Explanation Methods!LIME}%
\index{Explanation Methods!Surrogate Models}%
\index{Optimization-Based Explanations!Metamodels}%
\index{Surrogate-Based Explanations!LIME}
This method samples in the neighborhood of the input of interest, evaluates the
neural network at these points, and tries to fit the surrogate function such that it approximates the function of interest.
If the input domain of the surrogate function is human-interpretable, then LIME can even explain decisions of a model which uses non-interpretable features.
Since LIME is model agnostic, it can be applied to any classifier, even without knowing its internals, e.g., architecture or weights of a neural network classifier.
One major drawback of LIME is its high computational complexity, e.g., for state-of-the-art models such as GoogleNet it requires several minutes for computing the explanation of a single prediction \cite{samek:LapPhD19}.

Similar to LIME which builds a model for locally approximating the function of interest, the SmoothGrad method \cite{samek:smilkov2017smoothgrad}\index{Gradient-Based Explanations!SmoothGrad} samples the neighborhood of the input to approximate the gradient. Also SmoothGrad does not leverage the internals of the model, however, it needs access to the gradients. Thus, it can also be regarded as a gradient-based explanation method.

\subsection{Explaining with Local Perturbations}
Another class of methods construct explanations by analyzing the model's response to local changes. This includes methods which utilize the gradient information as well as 
perturbation- and optimization-based approaches.

Explanation methods relying on the gradient of the function of interest \cite{samek:ancona}\index{Explanation Methods!Gradient-Based}\index{Gradient-Based Explanations}\index{Gradient-Based Explanations!Sensitivity Analysis} have a long history in machine learning. One example is the so-called Sensitivity Analysis (SA) \cite{samek:morch1995visualization,samek:DBLP:journals/jmlr/BaehrensSHKHM10,samek:DBLP:journals/corr/SimonyanVZ13}. Although being widely used as explanation methods, SA technically explains the change in prediction instead of the prediction itself. Furthermore, SA has been shown to suffer from fundamental problems such as gradient shattering and explanation discontinuities, and is therefore considered suboptimal for explanation of today's AI models \cite{samek:montavon2017methods}.
Variants of Sensitivity Analysis exist which tackle some of these problems by locally averaging the gradients \cite{samek:smilkov2017smoothgrad} or integrating them along a specific path\index{Explanation Methods!Gradient-Based}\index{Gradient-Based Explanations!Integrated Gradients} \cite{samek:sundararajan2017axiomatic}.

Perturbation-based explanation methods \cite{samek:DBLP:conf/eccv/ZeilerF14,samek:zintgraf2017visualizing,samek:fong2017interpretable} explicitly test the model's response to more general local perturbations. While the occlusion method of \cite{samek:DBLP:conf/eccv/ZeilerF14} measures the importance of input dimensions by masking parts of the input, the Prediction Difference Analysis (PDA)\index{Explanation Methods!Perturbation-Based}\index{Perturbation-Based Explanations!Prediction Difference Analysis} approach of \cite{samek:zintgraf2017visualizing} uses conditional sampling within the pixel neighborhood of an analyzed feature to effectively remove information. Both methods are model-agnostic, i.e., can be applied to any classifier, but are computationally not very efficient, because the function of interest (e.g., neural network) needs to be evaluated for all perturbations.

The meaningful perturbation\index{Explanation Methods!Perturbation-Based}\index{Perturbation-Based Explanations!Meaningful Perturbations}\index{Optimization-Based Explanations!Meaningful Perturbations}\index{Explanation Methods!Meaningful Perturbations}\index{Meaningful Perturbations} method of \cite{samek:fong2017interpretable,samek:fong} is another model-agnostic technique to explaining with local perturbations. It regards explanation as a meta prediction task and applies optimization to synthesize the maximally informative explanations. The idea to formulate explanation as an optimization problem is also used by other methods.\index{Explanation Methods!Optimization-Based}\index{Optimization-Based Explanations} For instance, the methods \cite{samek:DBLP:journals/corr/SimonyanVZ13,samek:yosinski2015understanding,samek:nguyen2016synthesizing}\index{Optimization-Based Explanations!Activation Maximization}\index{Explanation Methods!Activation Maximization}\index{Activation Maximization} aim to interpret what the model has learned by building prototypes that are representative of the learned concept. These prototypes are computed within the activation maximization framework by searching for an input pattern that produces a maximum desired model response. Conceptually, activation maximization \cite{samek:nguyen2016synthesizing} is similar to the meaningful perturbation approach of \cite{samek:fong2017interpretable}. While the latter finds a minimum perturbation of the data that makes $f(x)$ low, activation maximization finds a minimum perturbation of the gray image that makes $f(x)$ high. The costs of optimization can make these methods computationally very demanding.

\subsection{Propagation-Based Approaches (Leveraging Structure)}
\index{Explanation Methods!Propagation-Based}
\index{Propagation-Based Explanations}
Propagation-based approaches to explanation are not oblivious to the model which they explain, but rather integrate the internal structure of the model into the explanation process.

Layer-wise Relevance Propagation (LRP) \cite{samek:BachPLOS15,samek:montavon2019}
\index{Explanation Methods!Layer-Wise Relevance Propagation}%
\index{Propagation-Based Explanations!Layer-Wise Relevance Propagation}%
\index{Layer-Wise Relevance Propagation}%
is a propagation-based explanation framework, which is applicable to general neural network structures, including deep neural networks \cite{samek:binder2016layer}, LSTMs \cite{samek:ArrWASSA17,samek:arras2019}, and Fisher Vector classifiers \cite{samek:LapCVPR16}. LRP explains individual decisions of a model by propagating the prediction from the output to the input using local redistribution rules. The propagation process can be theoretically embedded in the deep Taylor decomposition framework \cite{samek:MonPR17}.\index{Layer-Wise Relevance Propagation!Deep Taylor Decomposition}\index{Deep Taylor Decomposition} More recently, LRP was extended to a wider set of machine learning models, e.g., in clustering \cite{samek:kauffmann2019} or anomaly detection \cite{samek:kauffmann2018towards}, by first transforming the model into a neural network (`neuralization')\index{Neuralization} and then applying LRP to explain its predictions. The leveraging of the model structure together with the use of appropriate (theoretically-motivated) propagation rules, enables LRP to deliver good explanations at very low computational cost (one forward and one backward pass). Furthermore, the generality of the LRP framework allows also to express other recently proposed explanation techniques, e.g., \cite{samek:shrikumar2016not,samek:DBLP:conf/eccv/ZhangLBSS16}.
Since LRP does not rely on gradients, it does not suffer from problems such as gradient shattering and explanation discontinuities \cite{samek:montavon2017methods}.

Other popular explanation methods leveraging the model's internal structure are Deconvolution \cite{samek:DBLP:conf/eccv/ZeilerF14}\index{Explanation Methods!Propagation-Based}\index{Propagation-Based Explanations!Deconvolution} and Guided Backprogagation \cite{samek:springenberg2014striving}.\index{Explanation Methods!Propagation-Based}\index{Propagation-Based Explanations!Guided Backprop} In contrast to LRP, these methods do not explain the prediction in the sense ``how much did the input feature contribute to the prediction'', but rather identify patterns in input space, that relate to the analyzed network output.

Many other explanation methods have been proposed in the literature which fall into the ``leveraging structure'' category.
Some of these methods use heuristics to guide the redistribution process  \cite{samek:selvaraju2017grad}, others incorporate an optimization step into the propagation process \cite{samek:kindermans2017learning}. The iNNvestigate toolbox \cite{samek:AlbArXiv18} provides an efficient implementation for many of these propagation-based explanation methods.\index{Propagation-Based Explanations!iNNvestigate Toolbox}

\subsection{Meta-Explanations}
\index{Explanation Methods!Meta-Explanations}
\index{Meta-Explanations}
Finally, individual explanations can be aggregated and analyzed to identify general patterns of classifier behavior.
A recently proposed method, spectral relevance analysis (SpRAy) \cite{samek:LapNCOMM19},\index{Meta-Explanations!SpRAy} computes such meta explanations by clustering individual heatmaps.
This approach allows to investigate the predictions strategies of the classifier on the whole dataset in a (semi-)automated manner and to systematically find weak points in models or training datasets.

Another type of meta-explanation aims to better understand the learned representations and to provide interpretations in terms of human-friendly concepts. For instance, the network dissection\index{Network Dissection} approach of \cite{samek:bau2017network,samek:zhou}\index{Quantifying Interpretability} evaluates the semantics of hidden units, i.e., quantify what concepts these neurons encode.
Other recent work \cite{samek:kim2017interpretability}\index{Meta-Explanations!TCAV}\index{Gradient-Based Explanations!TCAV} provides explanations in terms of user-defined concepts and tests to which degree these concepts are important for the prediction.

\section{Evaluating Quality of Explanations}
\label{samek:sec:quality}
\index{Evaluating Explanations}
The objective assessment of the quality of explanations is an active field of research. Many efforts have been made to define quality measures for heatmaps which explain individual predictions of an AI model. This section gives an overview over the proposed approaches.

\index{Evaluating Explanations!Perturbation Analysis}
A popular measure for heatmap quality is based on perturbation analysis \cite{samek:BachPLOS15,samek:SamTNNLS16,samek:ArrPLOS17}. The assumption of this evaluation metric is that the perturbation of relevant (according to the heatmap) input variables should lead to a steeper decline of the prediction score than the perturbation of input dimensions which are of lesser importance. Thus, the average decline of the prediction score after several rounds of perturbation (starting from the most relevant input variables) defines an objective measure of heatmap quality. If the explanation identifies the truly relevant input variables, then the decline should be large. The authors of \cite{samek:SamTNNLS16} recommend to use untargeted perturbations (e.g., uniform noise) to allow fair comparison of different explanation methods. 
Although being very popular, it is clear that perturbation analysis can not be the only criterion to evaluate explanation quality, because one could easily design explanations techniques which would directly optimize this criterion. Examples are occlusion methods which were used in \cite{samek:DBLP:conf/eccv/ZeilerF14,samek:Li:ArXiv2017}, however, they have been shown to be inferior (according to other quality criteria) to explanation techniques such as LRP \cite{samek:ArrArXiv19}.

\index{Evaluating Explanations!Pointing Game}
Other studies use the `pointing game'' \cite{samek:DBLP:conf/eccv/ZhangLBSS16} to evaluate the quality of a heatmap. The goal of this game is to evaluate the discriminativeness of the explanations for localizing target objects, i.e., it is compared if the most relevant point of the heatmap lies on the object of designated category. Thus, these measures assume that the AI model will focus most attention on the object of interest when classifying it, therefore this should be reflected in the explanation. However, this assumption may not always be true, e.g., ``Clever Hans'' predictors \cite{samek:LapNCOMM19} may rather focus on context than of the object itself, irrespectively of the explanation method used. Thus, their explanations would be evaluated as poor quality according to this measure although they truly visualize the model's prediction strategy.

Task specific evaluation schemes have also been proposed in the literature. For example, \cite{samek:Poerner:ACL2018} use the subject-verb agreement task to evaluate explanations of a NLP model. Here the model predicts a verb's number and the explanations verify if the most relevant word is indeed the correct subject or a noun with the predicted number. Other approaches to evaluation rely on human judgment \cite{samek:ribeiro2016should,samek:Nguyen:NAACL2018}. Such evaluation schemes relatively quickly become impractical if evaluating a larger number of explanations.

\index{Evaluating Explanations!Toy Task}
A recent study \cite{samek:ArrArXiv19} proposes to objectively evaluate explanation for sequential data using ground truth information in a toy task. The idea of this evaluation metric is to add or subtract two numbers within an input sequence and measure the correlation between the relevances assigned to the elements of the sequence and the two input numbers. If the model is able to accurately perform the addition and subtraction task, then it must focus on these two numbers (other numbers in the sequence are random) and this must be reflected in the explanation.

\index{Evaluating Explanations!Indirect Evaluation}
An alternative and indirect way to evaluate the quality of explanations is to use them for solving other tasks. The authors of \cite{samek:ArrPLOS17} build document-level representations from word-level explanations. The performance of these document-level representations (e.g., in a classification task) reflect the quality of the word-level explanations. Another work \cite{samek:JAM:RUDDER2018} uses explanation for reinforcement learning. Many other functionally-grounded evaluations \cite{samek:doshi2017towards} could be conceived such as using explanations for compressing or pruning the neural network or training student models in a teacher-student scenario.

Lastly, another promising approach to evaluate explanations is based on the fulfillment of a certain axioms \cite{samek:shapley1953value,samek:sundararajan2017axiomatic,samek:lundberg2017unified,samek:montavon2017methods,samek:montavonb}.\index{Evaluating Explanations!Axioms} Axioms are properties of an explanation that are considered to be necessary and should therefore be fulfilled. Proposed axioms include relevance conservation \cite{samek:montavon2017methods}, explanation continuity \cite{samek:montavon2017methods}, sensitivity \cite{samek:sundararajan2017axiomatic} and implementation invariance \cite{samek:sundararajan2017axiomatic}. In contrast to the other quality measures discussed in this section, the fulfillment or non-fulfillment of certain axioms can be often shown analytically, i.e., does not require empirical evaluations.

\section{Challenges and Open Questions}
\label{samek:sec:challenges}
\index{Explainable AI!Challenges}
Although significant progress has been made in the field of explainable AI in the last years, challenges still exist both on the methods and theory side as well as regarding the way explanations are used in practice. Researchers have already started working on some of these challenges, e.g., the objective evaluation of explanation quality or the use of explanations beyond visualization. Other open questions, especially those concerning the theory, are more fundamental and more time will be required to give satisfactory answers to them.

Explanation methods allow us to gain insights into the functioning of the AI model. Yet, these methods are still limited in several ways.
First, heatmaps computed with today's explanation methods visualize ``first-order'' information, i.e., they show which input features have been identified as being relevant for the prediction.
However, the relation between these features, e.g., whether they are important on their own or only whether they occur together, remains unclear. 
Understanding these relations is important in many applications, e.g., in the neurosciences such higher-order explanations could help us to identify groups of brain regions which act together when solving a specific task (brain networks) rather than just identifying important single voxels.

Another limitation is the low abstraction level of explanations. Heatmaps show that particular pixels are important without relating these relevance values to more abstract concepts such as the objects or the scene displayed in the image. Humans need to interpret the explanations to make sense them and to understand the model's behaviour. This interpretation step can be difficult and erroneous. Meta-explanations which aggregate evidence from these low-level heatmaps and explain the model's behaviour on a more abstract, more human understandable level, are desirable. Recently, first approaches to aggregate low-level explanations \cite{samek:LapNCOMM19} and quantify the semantics of neural representations \cite{samek:bau2017network} have been proposed.
The construction of more advanced meta-explanations is a rewarding topic for future research.

Since the recipient of explanations is ultimately the human user, the use of explanations in human-machine interaction is an important future research topic.
Some works (e.g., \cite{samek:lage2019evaluation}) have already started to investigate human factors in explainable AI. Constructing explanations with the right user focus, i.e., asking the right questions in the right way, is a prerequisite to successful human-machine interaction.
However, the optimization of explanations for optimal human usage is still a challenge which needs further study.

A theory of explainable AI, with a formal and universally agreed definition of what explanations are, is lacking. 
Some works made a first step towards this goal by developing mathematically well-founded explanation methods. For instance, the authors of \cite{samek:MonPR17} approach the explanation problem by integrating it into the theoretical framework of Taylor decomposition.
The axiomatic approaches \cite{samek:sundararajan2017axiomatic,samek:lundberg2017unified,samek:montavon2017methods} constitute another promising direction towards the goal of developing a general theory of explainable AI. 

Finally, the use of explanations beyond visualization is a wide open challenge. Future work will show how to integrate explanations into a larger optimization process in order to, e.g.,  improve the model's performance or reduce its complexity.\\\\
{\bf Acknowledgements.} This work was supported by the German Ministry for Education and Research as Berlin Big Data Centre (01IS14013A), Berlin Center for Machine Learning (01IS18037I) and TraMeExCo (01IS18056A). Partial funding by DFG is acknowledged (EXC 2046/1, project-ID: 390685689). This work was also supported by the Institute for Information \& Communications Technology Planning \& Evaluation (IITP) grant funded by the Korea government (No.\ 2017-0-00451, No.\ 2017-0-01779).

%
%
\bibliographystyle{splncs04}

\endgroup

\end{document}